\documentclass[conf]{new-aiaa}
\usepackage[utf8]{inputenc}

\usepackage{graphicx}
\usepackage{amsmath}
\usepackage[version=4]{mhchem}
\usepackage{siunitx}
\usepackage{longtable,tabularx}
\usepackage{algorithm}
\usepackage{algpseudocode}
\usepackage{siunitx}
\usepackage{subcaption}

\newcommand{\pluseq}{\mathrel{+}=}
\newcommand{\minuseq}{\mathrel{-}=}

\title{Implementing TD3 to train a Neural Network to fly a Quadcopter through an FPV Gate}

\author{Patrick J. Thomas\footnote{Graduate Student, Kevin T. Crofton Department of Aerospace \& Ocean Engineering, Blacksburg, VA, 24061}}

\author{Kevin Schroeder\footnote{Research Scientist, Mission Systems Division, National Security Institute, Blacksburg, VA, 24061}}

\author{Jonathan T. Black\footnote{Professor, Kevin T. Crofton Department of Aerospace and Ocean Engineering; Director, Mission Systems Division, National Security Institute; Co-Director, Center for Space Science and Engineering Research(Space@VT), Blacksburg, VA, 24061}}
\affil{Virginia Polytechnic Institute and State University, Blacksburg, VA, 24061}

\begin{document}

\maketitle

\begin{abstract}
Deep Reinforcement learning has shown to be a powerful tool for developing policies in environments where an optimal solution is unclear. In this paper, we attempt to apply Twin Delayed Deep Deterministic Policy Gradients to train a neural network to act as a velocity controller for a quadcopter. The quadcopter's objective is to quickly fly through a gate while avoiding crashing into the gate. We transfer our trained policy to the real world by deploying it on a quadcopter in a laboratory environment. Finally, we demonstrate that the trained policy is able to navigate the drone to the gate in the real world.
\end{abstract}

\section{Introduction}
Over the past few years, Reinforcement Learning has shown to have the capacity to train Deep Neural Networks to perform complex tasks. This paper investigates the use of a Deep Reinforcement Learning algorithm, Twin Delayed Deep Deterministic Policy Gradient, to learn a policy to fly a quadcopter through a First Person View(FPV) drone racing gate.

\begin{figure}[h]
\begin{center}
\includegraphics[width=0.5\linewidth]{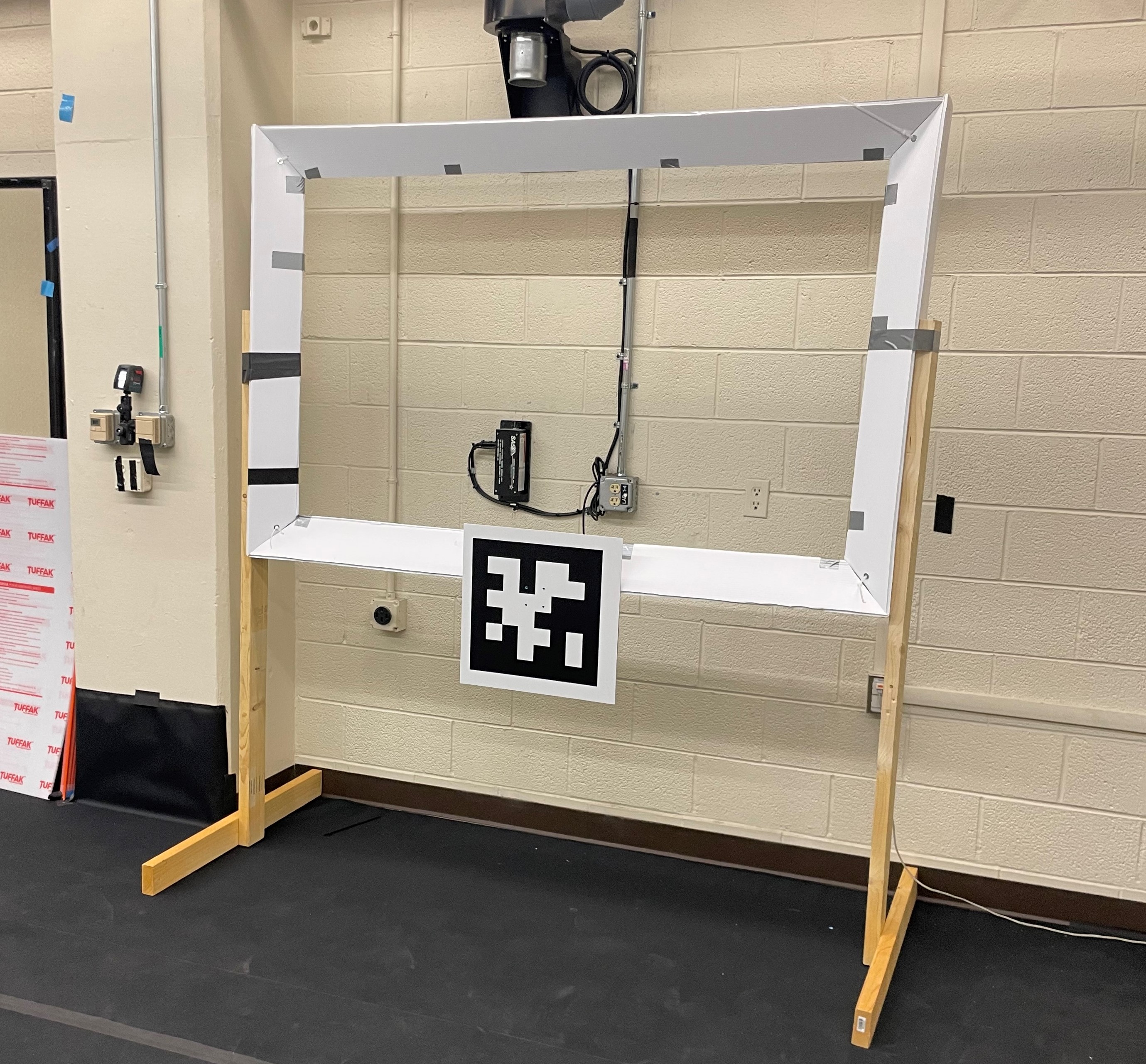}
\end{center}
   \caption{The gate design used for the BattleDrones competition. The AprilTag is located in the front lower center of the gate. The gate opening is 1.6 meters wide by 1 meter tall.}
\label{fig:gate}
\end{figure}

BattleDrones is an autonomous drone racing competition held by Virginia Tech. Teams must design a controller to navigate a quadcopter through a course consisting of multiple gates as part of the competition. The quadcopter is outfitted with a camera that is used to identify an AprilTag \cite{AprilTag}, a fiducial marker, on the gates. From this marker, the position and orientation of the gate relative to the drone is ascertained using a perspective-n-point algorithm. A gate is shown in Fig. \ref{fig:gate} for reference.

It is trivial to design a controller for the quadcopter that performs well while the vehicle operates in a deterministic environment. For example, a Proportional Derivative (PD) controller with reasonably chosen gains performs well at steering the drone to and through the gate in a deterministic environment. But, in a stochastic environment, the controller requires meticulous tuning to achieve good performance. The controller requires tuning as a result of the difficulties presented by flying through a gate with an uncertain position and orientation. Deep Reinforcement Learning is deployed deployed for these reasons using a simulated environment to train the controller. The controller is trained to determine velocity commands to safely and quickly fly a quadcopter to and through the center of a gate in a closed loop manner. 

The quadcopter is trained to fly to the origin of an environment to simplify the information the policy needs to learn. The gate is also placed normal to the East/Forward axis in and East-North-Up, Forward-Left-Up reference frame with its center at the origin removing the need to learn different gate orientations. The policy to train the drone is provided position and yaw relative to the Gate's normal as well as velocities rotated into the Gate's local coordinate frame for transferring the policy outside the simulated environment.

An environment is developed in gymnasium, the maintained version of OpenAI's Gym \cite{Gym}, to simulate the quadcopters motion. This environment is used to train a neural network that acts as a function approximator for the quadcopter's velocity controller. The Deep Reinforcement algorithm Twin Delayed Deep Deterministic Policy Gradients\cite{TD3} is deployed to train the neural network. The trained policy is tested by deploying it on a real world quadcopter. In real-world testing, the quadcopter attempts to fly through a virtual gate in a mixed reality setting.

\section{Methods}
\subsection{Gymnasium Environment}
A simulated environment is built that integrates with Gymnasium's existing architecture to facilitate the training of a controller that navigates a quadcopter through a gate. This simulated environment is created by following the \href{https://gymnasium.farama.org/tutorials/gymnasium_basics/environment_creation/}{guide} available on Gymnasium's website. The simulator uses a customs physics simulator to handle the vehicle dynamics. The physics simulator is built to model the simple motion of the quadcopter under given velocity commands, i.e. a point mass model. 

The controller is designed to use the drone's X, Y, and Z positions, yaw, X, Y, and Z velocities, and yaw rate as inputs. The controller outputs commanded X, Y, and Z velocities and a yaw rate. To meet these requirements, a simple simulator is created that models the quadcopter as a point that tracks it's pose and velocity as states. The simulator implements a first order model for the vehicle's response to a given velocity and yaw rate command:

\begin{equation} \label{eq:1}
    v_i(t) = \tau_i\frac{dv_i(t)}{dt},  i \in \{x, y, z, Yaw\}
\end{equation}

The response rate in the deterministic setting is modeled as $0.4 s$ for $V_x$ and $V_y$ and $0.1 s$ for $V_z$ and $V_{YAW}$. $V_{YAW}$ is the same as $\omega_z$ and used in place of $\omega_z$ in the rest of this paper. In the stochastic environment, $\tau$ is independently randomly sampled from a uniform distribution on the bounds $\{0.35 < \tau < 0.45\}$ for $V_X$ and $V_Y$ and on the bounds $\{0.08 < \tau < 0.13\}$ for $V_Z$ and $V_{YAW}$. A Tustin approximation of the transfer function described in equation \ref{eq:1} is applied to implement the model for velocity response in the discrete time simulator,

\begin{equation}
    V_{z+1, i} = aV_{cmd, z+1, i} + aV_{cmd, z, i} + bV_{z, i}
\end{equation}
\begin{equation}
    a = \frac{T_s}{2\tau_i + T_s}
\end{equation}
\begin{equation}
    b = \frac{2\tau - T_s}{2\tau + T_s}
\end{equation}
where $T_s$ is the period in seconds or $1/f_s$, the simulation frequency in Hertz. In the stochastic environment setting, noise is added to the velocity by sampling from a clipped normal.
\begin{equation}
    V_{k+1, i} = V_{k+1, i} + clip(\mathcal{N}(0, \sigma_i), min_i, max_i)
\end{equation}

Pose (X,Y,Z and Yaw) is determined by integrating the velocity. A zero order discrete time integrator is sufficient for the integrator because the simulation rate, $f_s$ is sufficiently high (50 Hz).

\begin{equation}
    P_{k+1} = P_k + V_{k+1}T_s
\end{equation}

In the stochastic environment setting, a randomly sampled clipped normal it added to position twice a second. This is implemented to model the changing estimate of where the gate is as the quadcopter gets more image information to estimate the gate's position.

\begin{equation}
    P_{k} = P{k} + clip(\mathcal{N}(0, \sigma_i), min_i, max_i)
\end{equation}

A human render mode is created for visualizing the quadcopter in the simulated environment. The open source python game engine \href{https://www.ursinaengine.org/}{Ursina} is used for rendering the environment visuals. The rendering provides visual feedback on the performance and movement of the quadcopter as it is commanded by a policy in the simulated environment. A snapshot from the environment is shown below in Fig. \ref{fig:ursina}.

\begin{figure}[h!]
\begin{center}
\includegraphics[width=0.5\linewidth]{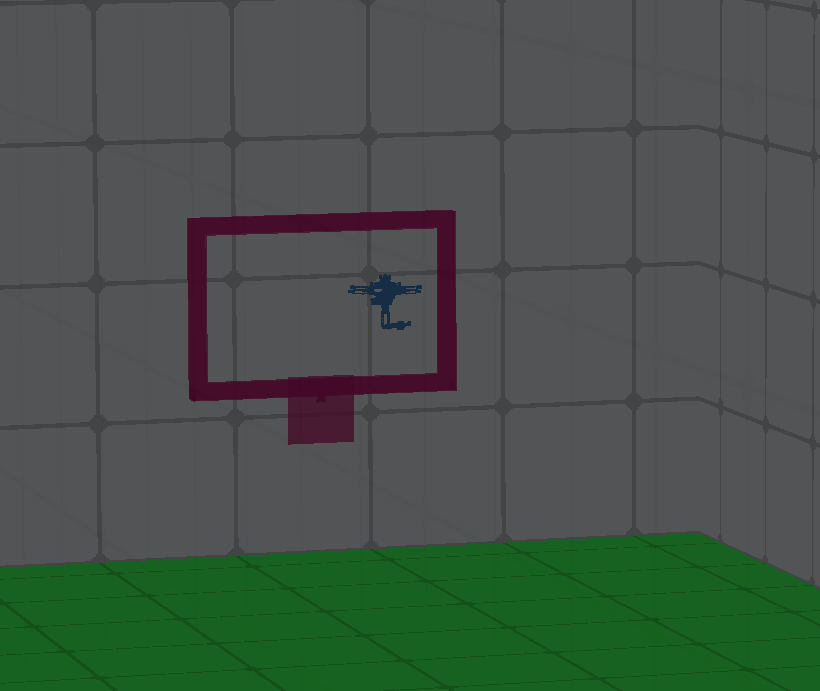}
\end{center}
   \caption{The drone and gate in the visualization of the simulated environment. The drone is the blue object and the gate is the maroon object. The bounds are shown by the gray walls and teh ground is shown by the green floor.}
\label{fig:ursina}
\end{figure}

The limits for the environment follow a box pattern. The X, Y, and Z position limits are [-10, 2], [-3, 3], and [-1.5, 2.5] $m$. Velocities are limited to [-2, 2], [-2, 2], [-1, 1] $m/s$, and [$-\frac{\pi}{2}, \frac{\pi}{2}$] $rad/s$ for the X, Y, Z, and yaw respectively. The drone's interference box is [0.6, 0.6, 0.2] $m$ in the X, Y, and Z. The gate's external dimensions are [0.15, 1.8, 1.22] $m$ in the X, Y, and Z. The thickness of the gate's walls is 0.12 $m$. A collision between the drone and gate is determined by checking for overlap between their boxed spaces. The simulation also has a time out period of 40 seconds or 2000 timesteps at 50 $Hz$.

A dense reward function is chosen for interacting with the environment. The reward function is designed to be simple enough for the drone to learn while strongly encouraging the drone to fly to the center of the gate, i.e. the origin of the environment. Penalties in the reward function are chosen to encourage the drone to avoid going out of bounds or crashing into the gate or ground. The reward function is described in Algorithm \ref{alg:Reward}.

\begin{algorithm}[t]
\caption{Quadcopter Reward Function}\label{alg:Reward}
\begin{algorithmic}
\State Dense Reward:
    \State $R = \num{3e-4}(\frac{\pi}{4} - \|Yaw\|)$
    \If{$X<0$ and $V_x > 0$}
        \State $R \pluseq \num{4e-2}(1 - \frac{\lVert \{X, Y, Z\} \rVert_{2}^{2}}{15})$ \Comment Proximity Based
    \ElsIf{$X > 0$}
        \State $R \minuseq \num{5e-2}$
    \Else
        \State $R \minuseq \num{1e-2}$
    \EndIf
\State Final Reward:
    \If{Gate contains Drone}
        \If{Gate Success Region contains Drone} \Comment{Success!}
            \State $R \pluseq 100$
            \State $R \pluseq 200(100^{-\lVert \{Y,Z\} \rVert_{2}^{2}})$
            \If{$\|Yaw\| < \frac{\pi}{6}$}
                \State $R \pluseq 100(1 - 3\frac{\|Yaw\|}{\pi})$
            \EndIf
        \Else \Comment{Crashed into Gate}
            \State $R \minuseq 20$
        \EndIf
    \EndIf
    \If{$Z < Z_{min}$} \Comment{Crashed into Ground}
        \State $R \minuseq 20$
    \EndIf
    \If {Simulation Space does not contains Drone} \Comment{Crashed into boundary}
        \State $R \minuseq 5$
    \EndIf
\end{algorithmic}
\end{algorithm}

\subsection{Twin Delayed Deep Deterministic Policy Gradient}
The action space for the quadcopter is continuous and therefore narrows the field of possible algorithms to train the policy. The algorithm implemented to train a neural network policy is Twin Delayed Deep Deterministic Policy Gradient (TD3) algorithm \cite{TD3}. Trust Region Policy Optimization (TRPO), Proximal Policy Optimization (PPO), and Deep Deterministic Policy Gradient (DDPG).\cite{TRPO}\cite{PPO}\cite{DDPG} are other algorithms that also allow training a policy for a continuous action space. These algorithms, while considered, are not implemented in this paper. TRPO and PPO are both on-policy algorithms. On-policy algorithms train on data collected from the current policy only, whereas DDPG and TD3 are off-policy algorithms. Off-policy algorithms make use of a replay buffer and can therefore be trained on data collected from roll outs of an earlier version of the policy. 

TD3 was introduced in 2018 by researchers from Canada and the Netherlands. This algorithm is designed to improve on the short comings of the DDPG algorithm. TD3 does this by incorporating a few key improvements. One such improvement is using clipped double-Q learning. That is, this algorithm learns two Q functions instead of just one and uses the smaller of the two in the Bellman error loss function. Another trick this algorithm uses is delayed policy updates. I.e., the algorithm trains the actor (policy) neural network less frequently than the critic (Q) network. TD3 is implemented in the same way as the original paper and updates the policy every other time step and updates the Q networks every time step. The third trick this algorithm employs is to stabilize the target policy by adding clipped normal noise to the target action. This makes it harder for the policy to exploit Q-function errors in the training. The TD3 algorithm implemented is shown in Algorithm \ref{alg:TD3}

\begin{algorithm}[t!]
\caption{Twin Delayed Deep Deterministic Policy Gradient}\label{alg:TD3}
\begin{algorithmic}
\State Input: Initial policy parameters $\theta$, Q-function parameters $\phi_1, \phi_2$, empty replay buffer $\mathcal{D}$
\State Set the target parameters equal to main parameters $\theta_{targ} \gets \theta$, $\phi_{targ,1} \gets \phi_1$, $\phi_{targ,2} \gets \phi_2$
\Repeat
    \State Observe state $s$ and select action $a=clip(\mu_{\theta}(s) + \epsilon, a_{Low}, a_{High})$, where $\epsilon \sim \mathcal{N}$
    \Comment{$\mathcal{N}$ is an Ornstein-Uhlenbeck process}
    \State Execute $a$ in the quadcopter environment
    \State Observe next state $s'$ and terminated signal $d$ to indicate whether $s'$ is terminal.
    \State Store $(s, a, r, s', d)$ in replay buffer $\mathcal{D}$
    \If {$s'$ is terminal or truncated}
        \State Reset environment state.
    \EndIf
    \If {it's time to update}
        \State Randomly sample a batch of transitions, $\mathcal{B} = \{(s, a, r, s', d\})$ from $\mathcal{D}$
        \State Compute target actions:
        \begin{align*}
            a'(s') = clip(\mu_{\theta_{targ}}(s') + clip(\epsilon, -c,c), a_{Low}, a_{High}), &\\   \epsilon \sim \mathcal{N}(0, \sigma)
        \end{align*}
        \State Compute targets
        \begin{align*}
            y(r,s',d) = r + \gamma(1-d)min_{i=1,2}Q_{\phi_{targ,i}}(s',a'(s'))
        \end{align*}
        \State Update Q-functions by one step of gradient descent using
        \begin{align*}
            \nabla_{\phi_{i}} \frac{1}{|B|_{(s,a,r,s',d)\in \mathcal{B}}} \sum (Q_{\phi_i}(s,a) - y(r,s',d))^2, &\\ \text{for }i=1,2
        \end{align*}
        \If{(j \% 2) == 0}
            \State Update policy by one step of gradient ascent using:
            \begin{align*}
                \nabla_{\theta} \frac{1}{|B|_{(s\in B)}} \sum{} Q_{\phi_1}(s,\mu_{\theta}(s))
            \end{align*}
            \State Update target network parameters with
            \begin{align*}
                \phi_{targ,i} \gets \rho \phi_{targ,i} + (1-\rho)\phi_i, &\\ \text{for }i = 1,2 &\\
                \theta_{targ} \gets \rho \theta_{targ} + (1-\rho)\theta
            \end{align*}
        \EndIf
    \EndIf
\Until{convergence}
\end{algorithmic}
\end{algorithm}

An Ornstein-Uhlenbeck process noise is added to the action selected by the policy. This noise process is draws from the DDPG process noise implementation\cite{DDPG}. This process noise is added to the policy action to improve exploration and prevent convergence to a sub-optimal solution. The Ornstein-Uhlenbeck process is a stationary Gauss-Markov process and can be considered a modification of the random walk in continuous time. Sigma and theta are chosen to be 0.15 and 0.2, respectively, the same values in the original DDPG algorithm.

\subsection{Real World Deployment}
The trained policy is deployed on a real world quadcopter the simulation was modeled after. The policy is implemented using TensorFlow Lite, a lightweight tool to run neural networks on edge processors. The quadcopter uses a Raspberry Pi 4B as an offboard computer that integrates with the flight controller. The trained policy is deployed directly on the quadcopter on the Raspberry Pi.

\begin{figure}[h!]
    \centering
    \includegraphics[width=0.5\linewidth]{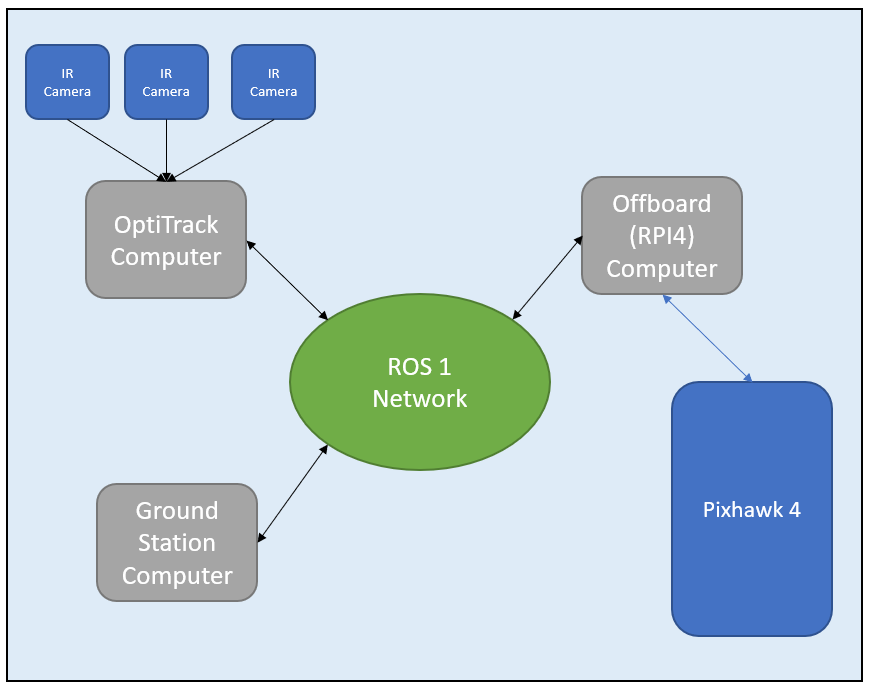}
    \caption{Diagram depicting the computer setup in the SpaceDrones lab. We use a distributed system with ROS 1 as the backbone for communicating between nodes on the Network.}
    \label{fig:lab_setup}
\end{figure}

The lab uses a distributed computing setup described by Gargioni et. al\cite{SpaceDrones} to facilitate real-world testing. The quadcopter uses a Pixhawk 4 running Px4 v1.13 firmware. The offboard computer is colocated with the Pixhawk 4 on the quadopter and connects to the network using ROS 1. The Raspberry Pi runs mavros, a ROS 1 package, that communicates with the flight controller in order to send position/orientation/velocity commands. In the lab environment, the quadcopter has its pose and orientation estimation supplemented with a ground truth value from an OptiTrack position system. The OptiTrack system uses a suite of Infrared (IR) cameras located around the lab and retro-reflective marker balls on the drone to determine its pose and orientation. A diagram of the lab setup is depicted in Fig. \ref{fig:lab_setup}

\section{Experimental Results}

\subsection{Training Results}
The final model used for the policy is a Vanilla Neural Network with 2 hidden layers. The model has an input size of 8, the first hidden layer has 400 nodes, the second hidden layer has 300 nodes, and the output layer has 4 nodes for the 4 velocity commands. The size of the hidden layers is chosen to follow the model that was used in the TD3 paper\cite{TD3}. The activation function chosen for the two hidden layers is Rectified Linear Unit (ReLu) and for the output layer is hyperbolic tangent (tanh). Using tanh as the activation function on the output layer ensures an output $\in (-1, 1)$ that is linearly scaled to the full scale range of the velocity commands for implementation. The networks are initialized using glorot normals for all nodes except the output nodes of the actor networks. For the actor networks, the output nodes are initialized by sampling from a uniform distribution $\in [-\num{1e-3}, \num{1e-3}]$. Keeping the output of the policy very small helped stabilize the initial training of the network.

The main hyperparameters that are adjusted during tuning are $\alpha$, the learning rates for the policy, $\beta$, the learning rate for the Q-functions, $\rho$, the delayed update rate for the target networks, and $\sigma$, the target noise variance. The hyperparameters and reward function are adjusted many times before a solution that would learn an effective policy is found. Because TD3 is a temporal difference learning algorithm, it can be very sensitive to learning rates. The initial learning parameters used for training came from the experiments documented in the TD3 paper and are tweaked from there. 




After a non-extensive hyper-parameter search, it is found that a value of $\alpha=\num{1e-5}$, $\beta=\num{2e-5}$, and $\tau=\num{1e-3}$ perform best. For the target process noise variance, $\sigma$, it is found that 0.2 performed well.

\begin{figure}[h!]
    \centering
    \includegraphics[width=0.7\linewidth]{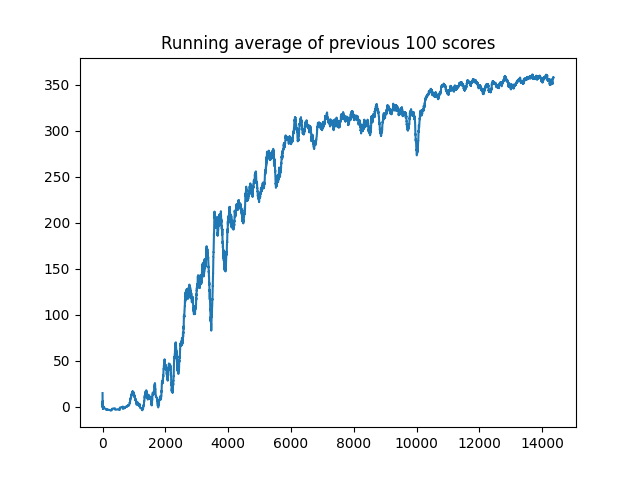}
    \caption{Training plot on the final training session. Learning across $\num{2.5e6}$ time steps. This amounts to about 15000 'games' played. A perfect score would be slightly higher than 400.}
    \label{fig:stable_learning}
\end{figure}

TD3 showed to be a very sample inefficient algorithm. In the training results, training typically took a minimum of $\sim \num 500e3$ steps to reach a point of learning a policy that would successfully fly through the gate. A plot of the policy network's learning is shown in Fig \ref{fig:stable_learning}. And further, it takes at least $\num{1.5e6}$ simulation steps to learn a policy that was excellent at reaching the gate while facing towards the gate along the way. Also of note, the simulation environment is not effected by the yaw of the drone, but, in the real world, the drone must be facing the gate in order for the camera to see the gate and thus be able to estimate the gate's position and orientation from the camera's images.

After training is completed, the policy is rolled out in the environment for 10 games to be visually inspected for it's performance.

\subsection{Real World Evaluation}

Finally, the trained policy is evaluated on one of our quadcopters in our lab. The test uses a mixed reality setup where the drone fly's in the real world while interacting with a virtual one. The drone's movement in the open and clear lab-space is used like a virtual reality (VR) tracker to move the virtual drone in the virtual environment. A virtual gate is placed in the virtual lab environment for the drone to interact with. The virtual drone has a virtual camera that is used to determine the location of the gate. The virtual world set up is created using Unreal Engine and was created by our colleague Minzhen Du.

\begin{figure}[h!]
    \centering
    \begin{subfigure}{0.45\linewidth}
    \centering
    \includegraphics[width=0.95\linewidth]{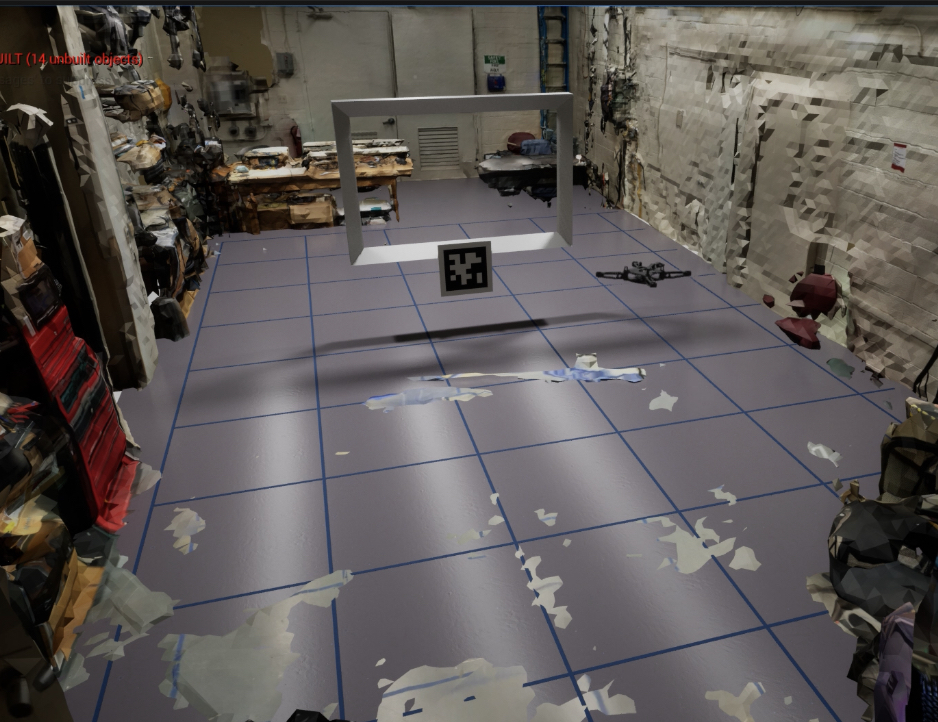} 
    \caption{Virtual World}
    \label{fig:subim1}
    \end{subfigure}
    \begin{subfigure}{0.45\linewidth}
    \centering
    \includegraphics[width=0.95\linewidth]{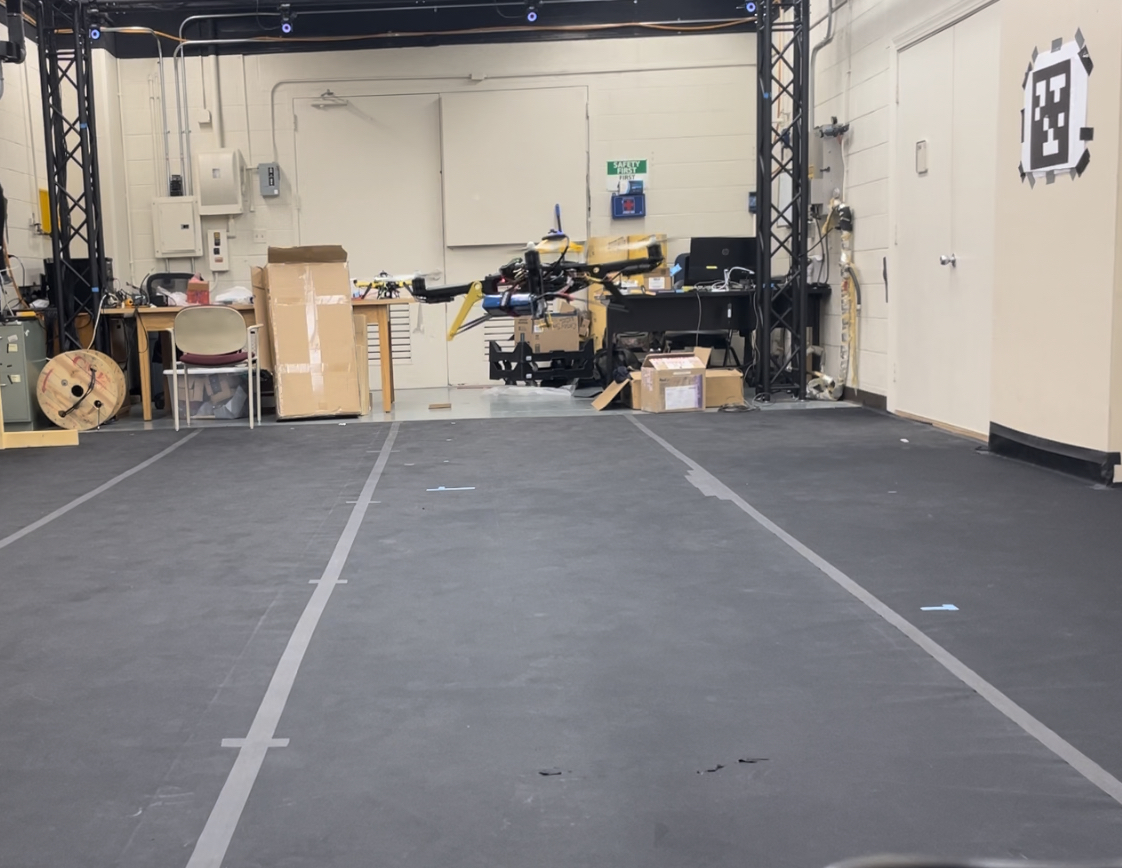}
    \caption{Real World}
    \label{fig:subim2}
    \end{subfigure}
    \caption{Images from the flight test showing the mixed reality setup. In Fig \ref{fig:subim1}, the drone in the virtual world can be seen with the virtual gate. Fig \ref{fig:subim2} shows the actual drone flying in an open lab space. The drones position and orientation in the real world is used to move the virtual drone in the simulated world.}
    \label{fig:Drone_flight}
\end{figure}

\begin{figure}[h!]
    \centering
    \begin{subfigure}[b]{0.45\linewidth}
    \centering
    \includegraphics[width=0.95\linewidth]{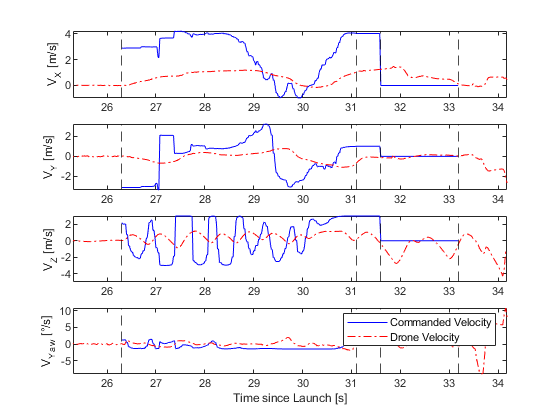} 
    \caption{Velocity Data}
    \label{fig:subim3}
    \end{subfigure}
    \begin{subfigure}[b]{0.45\linewidth}
    \centering
    \includegraphics[width=0.95\linewidth]{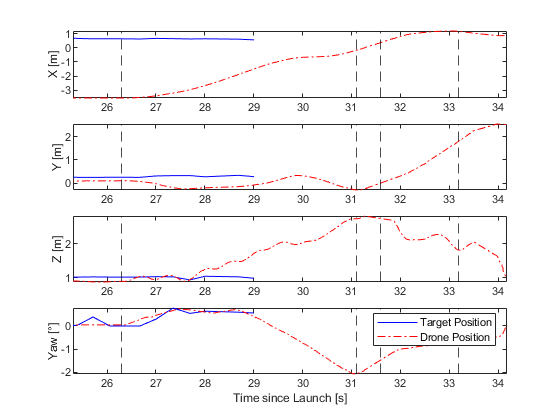}
    \caption{Position Data}
    \label{fig:subim4}
    \end{subfigure}
    \caption{Fig \ref{fig:subim3} shows the drones velocity and commanded velocity from the trained network during the flight test. Fig \ref{fig:subim4} shows the position and target position of the drone based on the most recent camera image during the flight test. The drone is overreacting in the Z direction likely because the vehicle response it was trained on was slower than actual. The position data shows the drone was attempting to move to the gate's center. The first vertical line indicates when the neural network was given control of the drone. The second vertical line indicates when the drone was determined to have made it to the gate in the X direction. The third vertical line indicates when all command velocities were set to 0 to stop the drone. And the fourth vertical line indicates when the drone was commanded to return home.}
    \label{fig:flight_test_1}
\end{figure}

\begin{figure}[h!]
    \centering
    \begin{subfigure}{0.45\linewidth}
        \centering
        \includegraphics[width=0.95\linewidth]{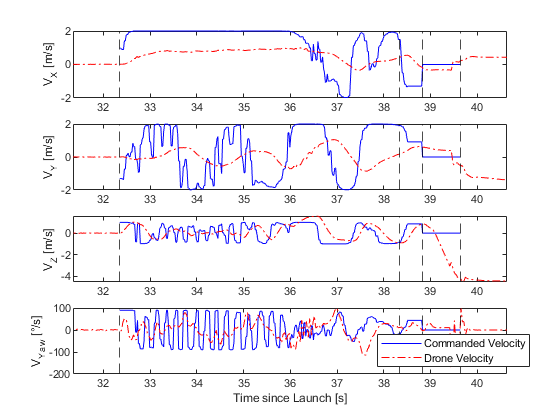} 
        \caption{Velocity Data}
        \label{fig:subim5}
    \end{subfigure}
    \begin{subfigure}{0.45\linewidth}
        \centering
        \includegraphics[width=0.95\linewidth]{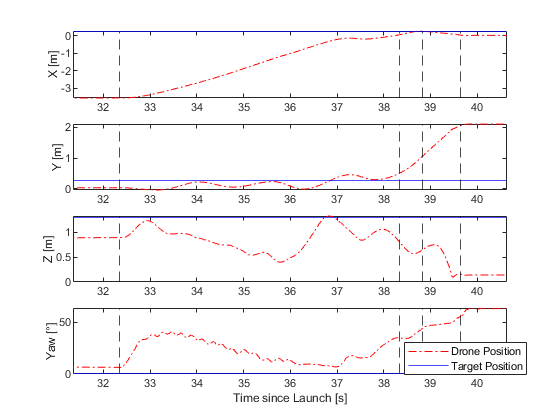}
        \caption{Position Data}
        \label{fig:subim6}
    \end{subfigure}
    \caption{Results from the second flight test using the network trained as shown in Fig \ref{fig:stable_learning}. Fig \ref{fig:subim5} shows the drones velocity and commanded velocity from the trained network during the flight test. Fig \ref{fig:subim6} shows the position and target position of the drone based on the most recent camera image during the flight test. Compared with the first flight attempt results in Fig \ref{fig:flight_test_1}, this new trained network performed much better.}
    \label{fig:flight_test_2}
\end{figure}

In testing in the real world, the drone showed to perform fairly poorly. Results from the test are plotted below in Fig \ref{fig:flight_test_1}. The policy that was run during this flight had been trained using an environment that modeled the vehicles response to input as all equal and sampled from a uniform distribution $\in[0.08, 0.2]s$. Further testing with the quadcopter, looking at its response to a step input, showed our model of the vehicles response was incorrect. It showed that the response to $V_z$ and $V_{YAW}$ commands was closer to $0.1s$ and the response for commanded $V_X$ and $V_Y$ was approximately $0.4s$. The network is retrained using the modified quadcopter model for $\num{2.5e6}$ time steps and then is deployed in the real world using the same mixed reality setup described above. Results from this second flight are shown in Fig. \ref{fig:flight_test_2}. Video of the flights is also available \href{https://www.youtube.com/watch?v=FD7yc-i9yu0}{here} for a visual comparison of performance. 

\section{Discussion}

While we were able to learn a policy for flying to the gate using TD3 in a simulated environment, the policy ultimately did not transfer well to the real world. This could be do to modelling errors of the quadcopter's dynamics in the simulated environment. We also neglected to include acceleration as an input to our network thus ensuring the system was a partially observable Markov decision process (POMDP) and thus a vanilla neural network is not necessarily the best choice. We think the results could be improved by modelling and including vehicle accelerations as part of the observable state space. We also wonder if incorporating recursive elements into the networks design like long-short term memory (LSTM) blocks as well as making the network deeper with more hidden layers could improve the results, especially when transferring the policy to the real world. 

Also, we think it would be interesting to incorporate model-based learning into the training process. We wonder if results can be improved by used a model of the environment that is pre-trained using the policies interaction with the simulated environment and then trained on data collected from the policy being rolled out in the real world. Finally, We think TD3 showed to be a tricky to use algorithm that easily became unstable and we would like to see if training performance improves using on-policy learning like PPO or TRPO. We wonder if the long roll-outs per episode are part of the reason temporal difference learning doesn't perform great. 



\section{Conclusions}

Ultimately, we were able to train a network to learn an effective policy for flying to the gate in the simulated environment. TD3 showed to be capable of training the policy using relatively simple reward functions. We were able to transfer our trained policy to the real world and run it on a real quadcopter using a Raspberry Pi 4B. The final trained policy, showed to be able to command the real world drone to gate center objective. While still far less stable than a simple PD controller, we believe we our on the right path to developing a successful controller using deep reinforcement learning in a simulated environment space and then transferring the trained policy to the real world.

{\small
\bibliography{report_bib}

\begin{thebibliography}{7}
\newcommand{\enquote}[1]{``#1''}
\providecommand{\natexlab}[1]{#1}
\providecommand{\url}[1]{\texttt{#1}}
\providecommand{\urlprefix}{URL }
\expandafter\ifx\csname urlstyle\endcsname\relax
  \providecommand{\doi}[1]{\discretionary{}{}{}https://doi.org/#1}\else
  \providecommand{\doi}[1]{\discretionary{}{}{}\urlstyle{rm}\url{https://doi.org/#1}}\fi

\bibitem[{Wang and Olson(2016)}]{AprilTag}
Wang, J., and Olson, E., \enquote{AprilTag 2: Efficient and robust fiducial
  detection,} \emph{2016 IEEE/RSJ International Conference on Intelligent
  Robots and Systems (IROS)}, 2016, pp. 4193--4198.
\newblock \doi{10.1109/IROS.2016.7759617}.

\bibitem[{Brockman et~al.(2016)Brockman, Cheung, Pettersson, Schneider,
  Schulman, Tang, and Zaremba}]{Gym}
Brockman, G., Cheung, V., Pettersson, L., Schneider, J., Schulman, J., Tang,
  J., and Zaremba, W., \enquote{Openai gym,} \emph{arXiv preprint
  arXiv:1606.01540}, 2016.

\bibitem[{Fujimoto et~al.(2018)Fujimoto, van Hoof, and Meger}]{TD3}
Fujimoto, S., van Hoof, H., and Meger, D., \enquote{Addressing Function
  Approximation Error in Actor-Critic Methods,} \emph{CoRR}, Vol.
  abs/1802.09477, 2018.
\newblock \urlprefix\url{http://arxiv.org/abs/1802.09477}.

\bibitem[{Schulman et~al.(2015)Schulman, Levine, Moritz, Jordan, and
  Abbeel}]{TRPO}
Schulman, J., Levine, S., Moritz, P., Jordan, M.~I., and Abbeel, P.,
  \enquote{Trust Region Policy Optimization,} \emph{CoRR}, Vol. abs/1502.05477,
  2015.
\newblock \urlprefix\url{http://arxiv.org/abs/1502.05477}.

\bibitem[{Schulman et~al.(2017)Schulman, Wolski, Dhariwal, Radford, and
  Klimov}]{PPO}
Schulman, J., Wolski, F., Dhariwal, P., Radford, A., and Klimov, O.,
  \enquote{Proximal Policy Optimization Algorithms,} \emph{CoRR}, Vol.
  abs/1707.06347, 2017.
\newblock \urlprefix\url{http://arxiv.org/abs/1707.06347}.

\bibitem[{Lillicrap et~al.(2015)Lillicrap, Hunt, Pritzel, Heess, Erez, Tassa,
  Silver, and Wierstra}]{DDPG}
Lillicrap, T.~P., Hunt, J.~J., Pritzel, A., Heess, N. M.~O., Erez, T., Tassa,
  Y., Silver, D., and Wierstra, D., \enquote{Continuous control with deep
  reinforcement learning,} \emph{CoRR}, Vol. abs/1509.02971, 2015.

\bibitem[{Gargioni et~al.(2019)Gargioni, Peterson, Persons, Schroeder, and
  Black}]{SpaceDrones}
Gargioni, G., Peterson, M., Persons, J.~B., Schroeder, K., and Black, J.,
  \enquote{A Full Distributed Multipurpose Autonomous Flight System Using 3D
  Position Tracking and ROS,} \emph{2019 International Conference on Unmanned
  Aircraft Systems (ICUAS)}, 2019, pp. 1458--1466.
\newblock \doi{10.1109/ICUAS.2019.8798163}.

\end{thebibliography}
}

\end{document}